\documentclass[10pt,twocolumn,letterpaper]{article}

\usepackage{iccv}
\usepackage{times}
\usepackage{epsfig}
\usepackage{graphicx}
\usepackage{amsmath}
\usepackage{amssymb}
\usepackage{booktabs}

\usepackage[breaklinks=true,bookmarks=false]{hyperref}

\iccvfinalcopy % *** Uncomment this line for the final submission

\def\iccvPaperID{****} % *** Enter the ICCV Paper ID here

\ificcvfinal\fi

\begin{document}

%%%%%%%%% TITLE
\title{3rd Place Solution to Large-scale Fine-grained Food Recognition}

\author{
Yang Zhong\\
East China Normal University\\
{\tt\small zhongyangtony@163.com}
\and
Yifan Yao\\
Shanghai Jiao Tong University\\
{\tt\small yao1010fan@sjtu.edu.cn}
\and
Tong Luo\\
OPPO Research Institute \\
{\tt\small luotong@oppo.com}
\and
Youcai Zhang\\
OPPO Research Institute\\
{\tt\small zhangyoucai@oppo.com}
\and
Yaqian Li\\
OPPO Research Institute \\
{\tt\small liyaqian@oppo.com}
}

\maketitle
\ificcvfinal\thispagestyle{empty}\fi

%%%%%%%%% ABSTRACT
\begin{abstract}
Food analysis is becoming a hot topic in health area, in which fine-grained food recognition task plays an important role. In this paper, we describe the details of our solution to the LargeFineFoodAI-ICCV Workshop-Recognition challenge\footnote{https://www.kaggle.com/c/largefinefoodai-iccv-recognition} held on Kaggle. We find a proper combination of Arcface loss\cite{deng2019arcface} and Circle loss\cite{sun2020circle} can bring improvement to the performance. With Arcface and the combined loss, 8 model was trained with carefully tuned configurations and ensembled to get the final results. Our solution won the 3rd place in the competition.
\end{abstract}

%%%%%%%%% BODY TEXT
\section{Introduction}
Large-scale fine-grained food recognition is a recognition competition focusing on classifying the fine-grained food images containing eastern and western food.

The provided large-scale food dataset\cite{Weiqing-LSVFR-CoRR2021} has 1500 food categories and about 80,0000 images\cite{Weiqing-LSVFR-CoRR2021}. Among them, 1000 food categories and 500,000 images, namely Food1K, are used for the food recognition task. Food1K is divided into train set, val set, and test set, containing 317277, 53462, and 158790 images respectively.

We found that adding Circle loss to Arcface can bring improvement, and models trained on different settings are helpful to ensembled results. Thus we trained our models on two sets of configurations, where different input size and loss functions are implemented. Distillation, test-time augmentation are also utilized and the final results are ensembled. With this pipeline, our results scored 0.92093 and 0.92019 on the public and private leaderboard respectively, winning the 3rd place of the competition.

% Our solution is based on a well-designed training strategy including two different configuration to train models, a combination of Arcface loss and Circle loss, distillation, test-time augmentation, and ensemble to enhance the models' performance further. Finally, it scored 0.92019 on the private leaderboard.

%Only those sampled from the train and val set are labeled. This recognition challenge asks us to feeds back the corresponding top-1 prediction results of each picture in the test set.

% This high-quality dataset is built through the initial food collection and finally evaluated by the professional annotators from the data annotation and quality inspection team in the Meituan company.

\section{Modeling}
To obtain food image representation, convolutional neural networks are employed through our pipeline. The backbones we used are RegNetX\cite{radosavovic2020designing}, EfficientNet-B5\cite{tan2019efficientnet}, EfficientNetV2\cite{tan2021efficientnetv2}, ResNeXt152\cite{xie2017aggregated}, ResNeSt200, and ResNeSt269\cite{zhang2020resnest}. We train all the models listed above with our training settings and distill two of them. Then we use test-time augmentation to improve their performance and use the ensemble as our final result.

\subsection{Basic Configuration}
Eight NVIDIA Tesla-V100 GPUs are used for training. During the training stage, we use two different configurations named config-a and config-b illustrated in Table \ref{tab:configs}. For config-b, we apply augmentations including random scale and center crop, horizontal flip with the probability of 0.5, 15-degree rotate, and color jitter\cite{krizhevsky2012imagenet} and set the training image size to 512x512. And for config-a, we additionally use cutmix\cite{yun2019cutmix} during augmentation and set the image size to 416x416 in training. The loss functions we use in config-a and config-b are respectively Arcface loss and the combined loss function(\ref{our_loss}). To get the optimal parameters, we tuned the settings on ResNeSt269 and transfer them to other backbones.
%As shown in Table \ref{tab:tricks}, ArcFace loss brings significant improvement. When adding Circle loss and remove cutmix in data augmentation, the model can achieve better score. 
%We tried config-a first on ResNeSt269 and achieve a score of 0.90510, then we made some experiments on a smaller model ResNeXt50\cite{xie2017aggregated} including changing image size, loss function, data augmentation, finetune strategy, and finally set config-b, which brought our score of ResNeSt269 to 0.91039. 

\begin{table}[htbp]
\begin{center}
\scalebox{0.95}{
\begin{tabular}{l c c c}
\toprule[1pt]
Configuration & config-a & config-b \\
\hline
Training size & 416 & 512 \\
Cutmix & \checkmark & \\
Optimizer & Momentum & Momentum \\
Scheduler & Cosine annealing & Cosine annealing \\
% Arcface loss & \checkmark & \checkmark \\
% Circle loss & & \checkmark \\
Testing size & 512 & 690 \\
\toprule[1pt]
\end{tabular}}
\end{center}
\caption{Configurations used in training and testing.}
\label{tab:configs}
\end{table}

\begin{table}[htbp]
\begin{center}
\scalebox{0.95}{
\begin{tabular}{l c c c}
\toprule[1pt]
Method & Input Size & Cutmix & Public \\
\hline
Cross Entropy & 416 & & 0.90195\\
Arcface loss & 416 & \checkmark & 0.90510 \\
Arcface loss & 512 & \checkmark & 0.90801 \\
Arcface loss + Circle loss & 512 & \checkmark & 0.90661 \\
Arcface loss + Circle loss & 512 & & 0.90892 \\
\toprule[1pt]
\end{tabular}}
\end{center}
\caption{Performance of ResNeSt269 with different configurations. }
\label{tab:tricks}
\end{table}

We apply config-a to ResNeSt200 and ResNeSt269 and denote them as rs200-a, rs269-a. Then we apply config-b to RegnetX-32gf, ResNeXt152, Efficient-V2, Efficient-B5, ResNeSt200 and ResNeSt269, which are named as regnetx, rx152, effnetv2, effnet5, rs200-b, rs269-b respectively in the following part.

We use different learning rates for different models in order to fit their training batch sizes and maximize the utility of GPUs. Our optimizer use SGD with momentum of 0.9. For the learning rate scheduler, cosine annealing\cite{loshchilov2016sgdr} is used and we set T as the max epoch in training.

\subsection{Backbone}
The aforementioned backbones are the most popular ones in recent years, apart from them, we also tried transformer frameworks, such as ViT\cite{dosovitskiy2020image} and CvT\cite{wu2021cvt}. However,  transformer-based models are not suitable in our case, since they rely on pretraining on a larger dataset to improve the performance. Besides, MLP-Mixer\cite{tolstikhin2021mlp} also underperformed. The performance of some backbones is listed in Table \ref{tab:backbone}. Inspired by \cite{touvron2019fixing}, we also implemented finetuning on a bigger image size, not only on the whole model but also on models of which the weights before the last batch normalization layer are frozen. However, both of them degraded the model's performance. 

\begin{table}[htbp]
\begin{center}
\begin{tabular}{l c c}
\toprule[1pt]
Backbone & Public & + TTA\\
\hline
CvT & 0.83787 & / \\
MLP-Mixer & 0.85978 & / \\
RegNetX & 0.87858 & 0.88742 \\
RegNetY & 0.87879 & / \\
EfficientNet-B5 & 0.89184 & 0.89430 \\
ResNeXt152 & 0.88880 & 0.89636\\
ResNeSt200-a & 0.89610 & / \\
ResNeSt200-b & 0.90708 & 0.90809\\
ResNeSt269-a & 0.90510 & 0.90833\\
ResNeSt269-b & 0.90892 & 0.91039\\
\toprule[1pt]
\end{tabular}
\end{center}
\caption{Single model results (TTA refers to test-time augmentation).}
\label{tab:backbone}
\end{table}

\subsection{loss Function}
To improve the performance, we tried several loss functions, including Cross Entropy loss, ArcFace loss\cite{deng2019arcface} and Circle loss\cite{sun2020circle}. Arcface performs better than the others, as it can enhance the discriminative power of fine-grained classification models. Then we tried the combination of different loss functions and found that adding Circle loss to ArcFace with a proper weight can accelerate convergence and further improve the performance of the model. 

We use ArcFace loss\cite{deng2019arcface} with a margin of 0.2 and a scale of 32 to train the models, then we use the Circle loss\cite{sun2020circle} with m of 0.25 and gamma of 32. In order to balance the role of the two functions in training, our loss is computed as: 
\begin{equation}\label{our_loss}
 \mathcal{L} = \gamma_{0}\mathcal{L}_{a} + \gamma_{1}\mathcal{L}_{c}
\end{equation}
where $\mathcal{L}$ is the loss function we use finally. $\mathcal{L}_{a}$ denotes ArcFace loss and $\mathcal{L}_{c}$ denotes Circle loss. $\gamma_{0}, \gamma_{1}$ are the weights of two loss functions, we set $\gamma_{0}$ as 1 and $\gamma_{1}$ as 1/$\beta$, where $\beta$ is the batch size used for training. During training, ArcFace loss plays a major role in the early stage and Circle loss works during the late stage. As shown in Table \ref{tab:tricks}, ArcFace loss brings significant improvement. When adding Circle loss and remove cutmix in data augmentation, the model can achieve better score. Thus the combined loss was used in config-a, and Arcface loss was utilized in config-b.
% Because after testing several sets of weights including setting the weights to 1:1, a combination of Cross Entropy loss and Circle loss, and using Circle loss only, we finally found that this combination of ArcFace loss and Circle loss is the best. ArcFace loss plays a major role in the early stage and Circle loss works during the later stage. 

\subsection{Distillation}
We tried BAKE\cite{ge2021self} first in our rs269-a, which is a self-distillation method that can be used while training, but the score was reduced to 0.90452 from 0.90510. Then We tried the traditional distillation method\cite{hinton2015distilling} with KDloss and cross entropy loss of equal weights and gained improvements. During distillation, KDloss is used to compute the loss of soft label, which is the 1000-dimension output embeddings of the teacher model and Cross Entropy loss is used to compute the loss of hard label, i.e., the ground truth. Considering the scale of our dataset and the way we distill, we set the temperature of KDloss to 3 and the weight of two loss functions to 1:1. Due to the limit of time and computation resources, we only distill two of the backbones, rx152 and rs200-b, with their trained versions.

\section{Post Processing}
Test-time augmentation is utilized to enhance the performance of single model, while ensemble method is used to integrate the output of different backbones. As shown in Table \ref{tab:results}, both of them make significant increase on the performance.

\begin{table}[htbp]
\begin{center}
\begin{tabular}{l c c}
\toprule[1pt]
Method & Public & Private\\
\hline
Single model & 0.90892 & 0.90760 \\
+ Test-time augmentation & 0.91039 & 0.90962 \\
+ Ensemble models & 0.92093 & 0.92019 \\
\toprule[1pt]
\end{tabular}
\end{center}
\caption{Performance of our pipeline on the public set and the private set from the leaderboard. Single model refers to ResNeSt269-b.}
\label{tab:results}
\end{table}

\subsection{Test-Time Augmentation}
In our pipeline, we utilize four kinds of test-time augmentations per test image, which are five crops, resize, horizontal flip, and rotate. This method can make up for the lack of information compared to a single input. We found that a slightly bigger input size leads to better performance, thus we use different input sizes during training and testing as in Table\ref{tab:configs}.

\subsection{Ensemble}
The ensemble method was used to further improve the performance. In our pipeline, final feature embeddings are the weighted sum of feature embeddings from the layer before the softmax of different models. The used models are RegNetX, EfficientNet-B5, EfficientNetV2, ResNeXt152, ResNeSt200-a, ResNeSt200-b, ResNeSt269-a, and ResNeSt269-b backbone models. Weights are given based on the performance of each model. 1.0 is assigned for the first five models, since the individual score of each model is below 0.90 on the test set. The weights of the rest three models are set to 1.5, as they can get a better score of nearly 0.91. As shown in Table \ref{tab:ensemble}, we also tried other ensemble methods including vote and stacking on RegNetY and ResNest269 in the early stage. However, they perform slightly worse than the sum of logits. An interesting phenomenon we found is that adding a model with the same structure from different code bases can bring improvement to the performance of the ensemble.

% \begin{table}[htbp]
% \begin{center}
% \scalebox{1.0}{
% \begin{tabular}{l c}
% \toprule[1pt]
% Backbone & Weight \\
% \hline
% regnetx & 1.0 \\
% effnet5 & 1.0 \\
% effnetv2 & 1.0 \\
% rx152 & 1.0 \\
% rs200-a & 1.5 \\
% rs200-b & 1.5 \\
% rs269-a & 1.5 \\
% rs269-b & 1.5 \\
% \toprule[1pt]
% \end{tabular}}
% \end{center}
% \caption{Assigned weight for each model.}
% \label{tab:weights}
% \end{table}

\begin{table}[htbp]
\begin{center}
\scalebox{1.0}{
\begin{tabular}{l c c}
\toprule[1pt]
Method & Public & Private \\
\hline
Vote & 0.91040 & 0.90960\\
Stacking & 0.91272 & 0.91209\\
Logits sum & 0.91339 & 0.91233 \\
\toprule[1pt]
\end{tabular}}
\end{center}
\caption{Different ensembled results of RegNetY and ResNeSt269.}
\label{tab:ensemble}
\end{table}

% \section{Results}
% We show the results of our pipeline in Table \ref{tab:results}. The score is significantly improved by the test-time augmentation and ensemble. In the LargeFineFoodAI-ICCV Workshop-Recognition challenge, our pipeline won 3rd place.

%------------------------------------------------------------------------
\section{Conclusion}
In this paper, we presented our approach for the LargeFineFoodAI-ICCV Workshop-Recognition challenge\footnote{https://www.kaggle.com/c/largefinefoodai-iccv-recognition} held on Kaggle in detail, which involves training strategy, distillation, test-time augmentation, and ensemble method. With careful parameter tuning, a new combination of Arcface loss and Circle loss and ensemble, we won 3rd place in the competition.
% In this paper, we presented a large-scale fine-grained food recognition pipeline by team OPPO Research Institute. Our experimental results show that our training strategy, distillation, test-time augmentation, and finally ensemble method can be combined together and achieve better results. Finally, we won the 3rd place in the LargeFineFoodAI-ICCV Workshop-Recognition challenge\footnote{https://www.kaggle.com/c/largefinefoodai-iccv-recognition} held on Kaggle.

{\small
\bibliographystyle{ieee_fullname}
\bibliography{egbib}
}

\end{document}